\documentclass[10pt,twocolumn,letterpaper]{article}

\usepackage{cvpr}
\usepackage{times}
\usepackage{epsfig}
\usepackage{graphicx}
\usepackage{amsmath}
\usepackage{amssymb}

\usepackage{float}

\newcommand{\vtext}[1]{\rotatebox[origin=b]{90}{#1}}

\usepackage{etoolbox}
\usepackage{bm}
\usepackage{amsthm}%
\usepackage{algorithm}%
\usepackage{algorithmic}%
\usepackage{subfigure}%

\newcommand{\MyMapTemplatePrefixc}[4]{\expandafter#1\csname#3#4\endcsname{#2{#4}}} %
\forcsvlist{\MyMapTemplatePrefixc {\def} {\mathcal}{c}} {A,B,C,D,E,F,G,H,I,J,K,L,M,N,O,P,Q,R,S,T,U,V,W,X,Y,Z}  %

\newcommand{\MyMapTemplatePrefixtb}[5]{\expandafter#1\csname#4#5\endcsname{#2{#3{#5}}}} %
\forcsvlist{\MyMapTemplatePrefixtb {\def} {\tilde}{\mathbf}{t}} {A,B,C,D,E,F,G,H,I,J,K,L,M,N,O,P,Q,R,S,T,U,V,W,X,Y,Z}  %
\forcsvlist{\MyMapTemplatePrefixtb {\def} {\tilde}{\mathbf}{t}} {0,1,a,b,c,d,e,f,g,h,i,j,k,l,m,n,o,p,q,r,s,u,v,w,x,y,z}  %

\newcommand{\MyMapTemplateNoPrefix}[3]{\expandafter#1\csname#3\endcsname{#2{#3}}}
\forcsvlist{\MyMapTemplateNoPrefix {\def} {\mathbf} } {0,1,a,b,c,d,e, f, g, h, i, j, k, l, m, n, o, p, q, r, u, v, w, x, y, z} %
\forcsvlist{\MyMapTemplateNoPrefix {\def} {\mathbf} } {A,B,C,D,E,F,G,H,I,J,K,L,M,N,O,P,Q,R,S,T,U,V,W,X,Y,Z}  %

\def\resp{\emph{resp.}\@\xspace}

\usepackage{booktabs} %
\usepackage[table,dvipsnames]{xcolor}
\definecolor{rowblue}{RGB}{220,230,240}

\usepackage[pagebackref=true,breaklinks=true,letterpaper=true,colorlinks,bookmarks=false]{hyperref}

\cvprfinalcopy %

\def\cvprPaperID{642} %

\ifcvprfinal\fi
\begin{document}

\title{Learning Semantic Segmentation from Synthetic Data: \\ A Geometrically Guided Input-Output Adaptation Approach}

\author{Yuhua Chen$^1$\hspace{10mm}Wen Li$^1$\hspace{10mm}Xiaoran Chen$^1$\hspace{10mm}Luc Van Gool$^{1,2}$\\[2mm]
$^1$Computer Vision Laboratory, ETH Zurich\hspace{10mm}
$^2$VISICS, ESAT/PSI, KU Leuven\\[-1.5pt]
{\tt\small \{yuhua.chen,liwen,chenx,vangool\}@vision.ee.ethz.ch}
}

\maketitle

\begin{abstract}
Recently, increasing attention has been drawn to training semantic segmentation models using synthetic data and computer-generated annotation. However, domain gap remains a major barrier and prevents models learned from synthetic data from generalizing well to real-world applications. In this work, we take the advantage of additional geometric information from synthetic data, a powerful yet largely neglected cue, to bridge the domain gap. Such geometric information can be generated easily from synthetic data, and is proven to be closely coupled with semantic information. With the geometric information, we propose a model to reduce domain shift on two levels: on the input level, we augment the traditional image translation network with the additional geometric information to translate synthetic images into realistic styles; on the output level, we build a task network which simultaneously performs depth estimation and semantic segmentation on the synthetic data. Meanwhile, we encourage the network to preserve the correlation between depth and semantics by adversarial training on the output space. We then validate our method on two pairs of synthetic to real dataset: Virtual KITTI$\rightarrow$KITTI, and SYNTHIA$\rightarrow$Cityscapes, where we achieve a significant performance gain compared to the non-adaptive baseline and methods without using geometric information. This demonstrates the usefulness of geometric information from synthetic data for cross-domain semantic segmentation.
\end{abstract}

\begin{figure}
  \center
  \includegraphics[trim={0cm 9cm 16.3cm 0},clip,width=\linewidth]{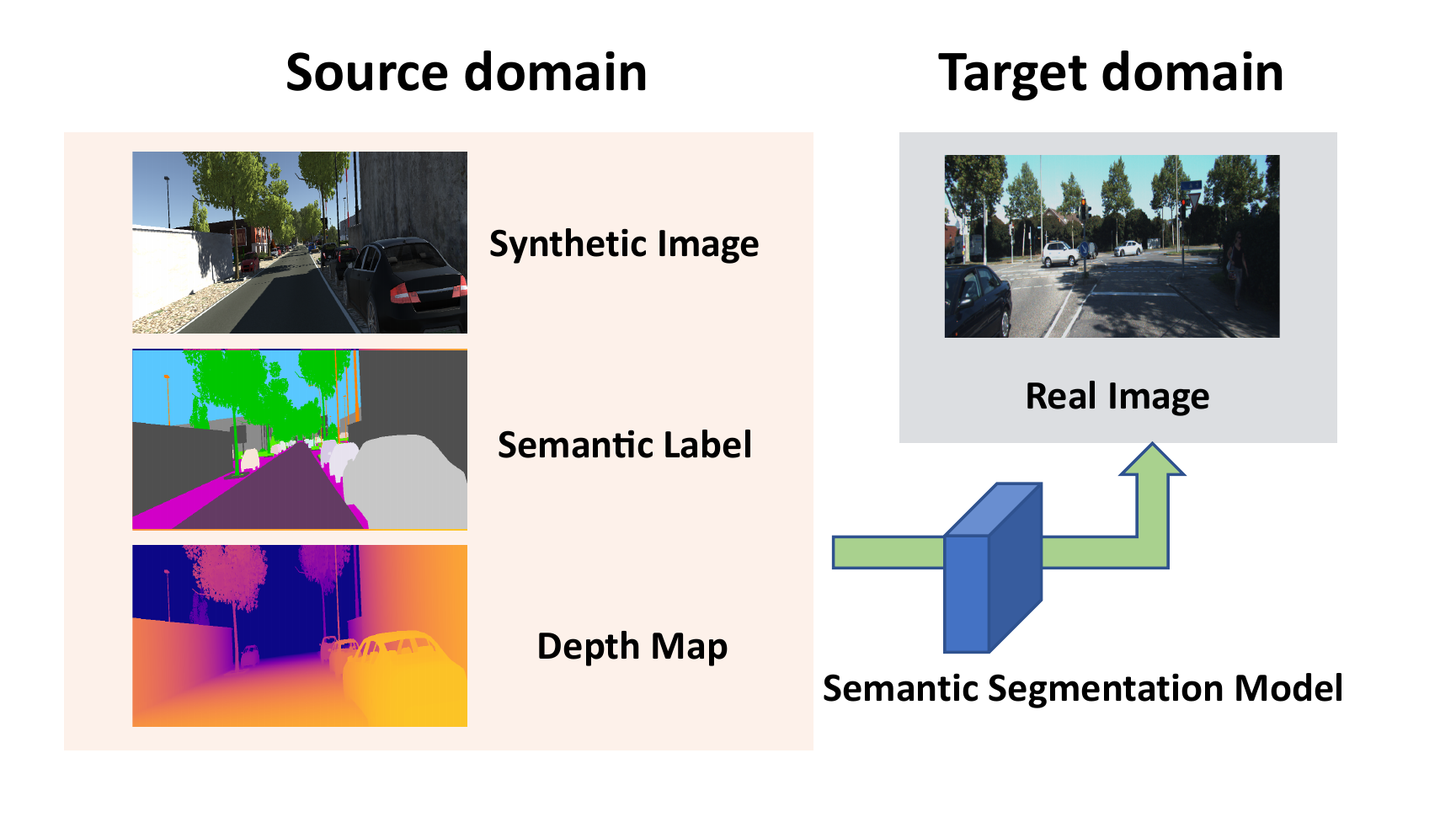}
  \caption{Our work aims to learn a semantic segmentation model from synthetic data, which can also be applied to real-world data. The domain adaptation is reinforced by geometric information in the synthetic data, which can be obtained easily from virtual environment.}
  \label{fig:setting}
  \vspace{-5mm}
\end{figure}

\section{Introduction}
Semantic segmentation of urban scenes refers to the task of assigning each pixel in a given scene with a semantic category, such as \textit{sky}, \textit{road}, \textit{car}, etc. There is a growing interest in this task in recent years~\cite{cordts2016cityscapes,zhao2016pyramid}, partly due to its many attractive applications, such as autonomous driving. Like many other tasks in computer vision, deep neural networks~\cite{krizhevsky2012imagenet} excel at semantic segmentation when trained on large-scale labeled datasets. However, building such datasets is not an easy task, in terms of both collection and annotation: it is non-trivial to collect images with large diversity of urban scenes for different weathers, cities and other conditions; labelling these images can be even more challenging due to the labor-intensive process of dense annotation for complicated urban scenes.

Due to these limitations, learning from synthetic data becomes a charming alternative to train semantic segmentation models. Recent advances in computer graphics make it possible to generate synthetic images, and per-pixel annotations from virtual 3D environments~\cite{richter2016playing,richter2017playing}. Training on synthetic data seems to be a tempting way to reduce manual labelling cost, however, the mismatch in appearance often leads to a significant performance drop when the learned models are applied to real environment. With that being the case, the key to effective utilization of synthetic data is to overcome such domain mismatch.

Many works address this issue from the perspective of domain distribution shift with various domain adaptation techniques~\cite{hoffman2016fcns,zhang2017curriculum,chen2018road}. On the other hand, image translation~\cite{zhu2017unpaired} has been widely deployed to transform the synthetic images into images of realistic styles. This can be seen as aligning the domain distribution at pixel level~\cite{hoffman2017cycada}. Nevertheless, these works typically utilize only synthetic images and their semantic labels. However, a significant advantage of the synthetic data has been unfortunately overlooked: one can actually obtain rich information from the virtual environment, including surface norm, optical flow, depth, etc., at a much lower cost than obtaining the same kinds of information in real-world data. 

As illustrated in Figure~\ref{fig:setting}, the aim of this work is to exploit the additional geometric information from the synthetic domain to improve cross-domain semantic segmentation in real data. As a matter of fact, geometry and semantics are two strongly correlated aspects of urban scenes as street scenes mostly follow similar layouts. Geometric cue can usually imply semantic information and vice verse. As shown in previous works~\cite{wang2015towards}, joint reasoning the depth and semantics improve the performance of both tasks. Additionally unlike the large gap between synthetic and real images, such correlation suffers less from domain shift. For example, road is usually flat, sky is far away, poles are vertical. These facts hold regardless in synthetic data and real data. Thus, the correlation between semantics and geometry is highly favoured for predicting semantics and reducing the domain gap. Besides, depth information is relatively easy to acquire from synthetic data, as one simply needs to generate it from virtual 3D model, and no special equipment (lidar, calibrated stereo cameras) is needed.

We present a new approach called \textit{Geometrically Guided Input-Output Adaptation} (GIO-Ada), in which we integrate depth information into our domain adaptation task on the following levels: 1) on the input level, an augmented image transform network takes synthetic image and its corresponding semantic and depth map as input, and is trained to produce images with realistic style by exploiting internal correlation of raw images, semantic and geometric information; and 2) on the output level, a task network jointly performs depth estimation and semantic segmentation using supervision from synthetic domain. Further, adversarial training is applied on the joint output space of semantic segmentation and depth estimation and preserves domain-invariant correlation between depth and semantics. With the above two modules, geometric information not only improves the prediction of semantics, but also help to alleviate the domain gap between synthetic and real data. 

The proposed framework is validated through extensive experiments on Virtual KITTI~\cite{gaidon2016virtual}, KITTI~\cite{geiger2013vision}, SYNTHIA~\cite{ros2016synthia}, and Cityscapes~\cite{cordts2016cityscapes} datasets, where we observe significant performance improvements compared to the non-adaptive baseline and other methods without using geometric information, which demonstrates the effectiveness of our model on incorporating geometric information for cross-domain semantic segmentation from synthetic data to real scenarios. 

\section{Related Works}
\textbf{Semantic Segmentation} is a highly active research field. Recent approaches are mostly based on fully convolutional network~\cite{long2015fully} with modifications designed for pixel-wise prediction, such as DilatedNet~\cite{yu2015multi}, DeepLab~\cite{chen2016deeplab}, PSPNet~\cite{zhao2016pyramid}~\etc.

Such models are generally trained on datasets with pixel-wise annotation, for example, PASCAL~\cite{everingham2010pascal}, COCO~\cite{lin2014microsoft}, and Cityscapes~\cite{cordts2016cityscapes} which is more related to urban scene scenarios. However, building such datasets and collecting per-pixel annotations are both expensive and laborious. With the development of computer graphics techniques, synthetic data enables an alternative approach to training semantic segmentation models at a lower cost. To this end, several synthetic datasets have been built, for example, GTAV~\cite{richter2016playing}, SYNTHIA~\cite{ros2016synthia}, Virtual KITTI~\cite{gaidon2016virtual}, etc. These datasets are typically generated from virtual 3D models, meaning that modalities other than the semantic label map can be generated easily. Such modalities include optical flow, depth, surface normal etc. Therefore, our work is motivated to leverage such free supervision signals in synthetic data in order to effectively perform cross-domain semantic segmentation.

\textbf{Domain Adaptation} is a classic problem in both machine learning and computer vision. It aims to mitigate the performance drop caused by the distribution mismatch between training and test data. It is mostly studied in image recognition problems by both conventional methods~\cite{kulis2011you,gopalan2011domain,gong2012geodesic,fernando2013unsupervised}, and CNN-based methods \cite{long2015learning,ganin2015unsupervised,ghifary2016deep,sener2016learning,panareda2017open,motiian2017unified,li2017deeper,haeusser2017associative,lu2017unsupervised,maria2017autodial}. We refer interested readers to \cite{patel2015visualdomainadaptation,csurka2017domain} for comprehensive surveys.

Our work is more related to cross-domain semantic segmentation~\cite{hoffman2016fcns,zhang2017curriculum,hoffman2017cycada,tsai2018learning,sankaranarayanan2017unsupervised,chen2018road,zou2018unsupervised,zhu2018penalizing}. The first work to investigate the cross-domain urban scene semantic segmentation is \cite{hoffman2016fcns}, where they deploy adversarial training to align the features from different domains.  Following this line, many works have been proposed to address the domain shift problem in semantic segmentation by different techniques, such as curriculum style learning~\cite{zhang2017curriculum}, cycle consistency~\cite{hoffman2017cycada}, output space alignment~\cite{tsai2018learning}, generative adversarial network~\cite{sankaranarayanan2017unsupervised}, distillation loss~\cite{chen2018road}, class-balanced self-training~\cite{zou2018unsupervised}, conservative loss~\cite{zhu2018penalizing}, etc. Moreover, inspired by the success of generative adversarial network~\cite{radford2015unsupervised,goodfellow2014generative} and image translation techniques~\cite{zhu2017unpaired,isola2017image}, a few works have also suggested to transform synthetic images with realistic styles to reduce domain gap on raw-pixel level~\cite{shrivastava2017learning,hoffman2017cycada,murez2017image} and to boost the semantic segmentation performance in real scenarios. However, the mentioned works typically only rely on labelled synthetic images and unlabelled real images while neglecting other free information in the dataset, such as geometric information.

\begin{figure*}[h]
  \center
  \includegraphics[trim={0cm 5cm 0cm 0},clip,width=0.95\textwidth]{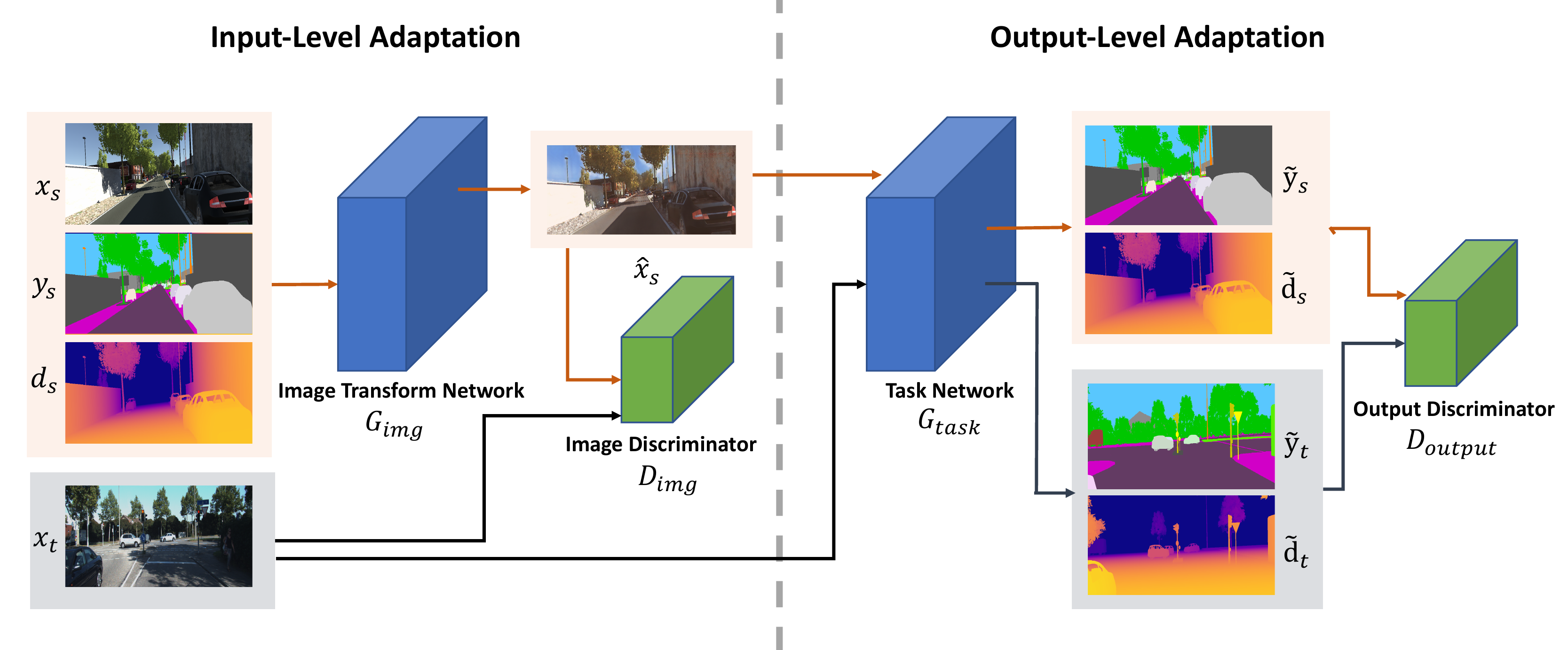}
  \caption{\textbf{Overview of the Proposed Architecture.} The flow of source data is shown in orange line, while the flow of target data is shown in black line.}
  \label{fig:architecture}
\end{figure*}

\textbf{Geometry Guided Semantic Segmentation:} Depth  estimation  and  semantic  segmentation are two fundamental aspects of scene understanding, as the two tasks are strongly correlated. There are many works proposed to jointly learn these two tasks in a mutually beneficial manner. For instance, \cite{wang2015towards} built a hierarchical CRF with CNN to leverage the geometric cue, and \cite{kendall2017multi} proposed a cross-task uncertainty. There are other works proposed to jointly learn the two tasks with various techniques, including fine-tuning~\cite{mousavian2016joint}, cross-modality influence~\cite{jafari2017analyzing}, task distillation module with intermediate auxiliary tasks~\cite{xu2018pad}, recursive estimation~\cite{zhang2018joint}, task attention loss~\cite{jiao2018look}. 

More broadly speaking, the idea of jointly learning semantic segmentation and depth estimation can be connected to multi-task learning~\cite{kokkinos2017ubernet}, where multiple outputs are produced by a single network. 

Different from previous research, our aim in this work is to leverage the correlation between depth and semantic segmentation, for the purpose of domain adaptation.

\section{Geometrically Guided Input-Output Adaptation}
In this work, we aim to learn the semantic segmentation model by leveraging synthetic data, and apply the learned model to real-world scenarios. Following unsupervised domain adaptation protocol, synthetic data is utilized as the source domain $S$, and real data as the target domain $T$. In the source domain, we have access to synthetic images $x_s\in S$ along with their corresponding ground-truth labels, including semantic segmentation labels ${y}_s$ and depth labels ${d}_s$. In the target domain, we only have access to unlabelled images $x_t\in T$. 

\subsection{Framework Overview}
The overview of the our proposed Geometrically Guided Input-Output Adaptation (GIO-Ada) approach is illustrated in Figure~\ref{fig:architecture}. To address the domain gap between synthetic and real domains, domain adaptation is performed jointly on two levels, namely input level and output level. Depth information (\ie, geometric information) is exploited for improving adaptation on both levels.

\textbf{Input-level adaptation} aims to reduce visual differences at raw pixel level. The output from input-level adaptation are later used as input to the followed task network. For this purpose, we deploy an image transform network $G_{img}$ which takes a synthetic image $x_s$, along with its corresponding depth ${d}_s$ and semantic label ${y}_s$ as input. The transform network $G_{img}$ is supposed to produce transformed images $\hat{x}_s$ with visually similar appearances to images in the target domain, and at the same time preserves useful information for semantic segmentation and depth estimation. 

Most of the existing pixel-level adaptation methods do not consider adding depth information to the source domain, and use only synthetic images as source data. This is apparently not an optimal way. When rendering synthetic images from 3D models, geometric information are mostly lost, and such information is difficult to recover once discarded. On the other hand, it has been shown in recent works \cite{zhang2018joint,xu2018pad} that geometric information is highly correlated to semantic information. Therefore, we incorporate it into our image transform network to better preserve the semantic information during image translation.

\textbf{Output-level adaptation} aims to align predictions of the task network for two domains, and also retain the coherent correlation between tasks. The output-level adaptation includes a task network $G_{task}$ and an output-level discriminator $D_{output}$. $G_{task}$ takes real images $x_t$ or transformed synthetic images $\hat{x}_s$ as input, then simultaneously predicts semantic segmentation $\widetilde{y}$ and depth prediction $\widetilde{d}$. $D_{output}$ tries to determine if the outputs (depth and semantics) are predicted from a transformed synthetic image or a real one. 

Utilizing geometric information in output-level adaptation brings several benefits. First, by learning depth estimation as an auxiliary task, we can learn representation which is more robust against domain shift. Second, the correlation between depth and semantics can be used as a powerful cue for domain alignment. In the target domain, since the model has no access to the annotations, therefore aligning the output space between the two domains can be a highly useful supervision signal to guide the training. Unlike previous works~\cite{tsai2018learning} which only aligns the output space of a single task, here we consider the joint output space of both depth estimation and semantic segmentation. In this way, we align not only the output distributions of each individual task, but also the underlying interconnection between different tasks. This is proven to be effective for boosting the performance of the two tasks. It is also consistent with our motivation that such connections suffer less from domain shift, for instance, sky is always far away, cars are usually on the street etc. Hereby, we respectively elaborate the adaptation on the two levels in the following sections.

\subsection{Input-Level Adaptation}
To transform synthetic images into the real-style images, we build an image transform network $G_{img}$ with synthetic image $x_s$, semantic segmentation map ${y}_s$ and depth map ${d}_s$ as input. In particular, the depth map is normalized into a range of $[0,1]$, and the semantic label map is represented as a one-hot map with $C$ channels where $C$ is the total number of categories. The network outputs the transformed image $\hat{x}_s = G_{img}(x_s,{y}_s,{d}_s)$, which is expected to be realistic-looking and still contains vital information for the task networks  (\eg, semantic segmentation,  depth estimation.).

Inspired by generative adversarial networks (GANs)~\cite{goodfellow2014generative}, we apply a discriminator $D_{img}$ to guarantee the realism of generated images. The discriminator $D_{img}$ is trained to distinguish between transformed synthetic images and real images. At the same time, $D_{img}$ is also used to guide the training of the image transform network in a similar way to the adversarial training strategy in GANs. Specifically, we use PatchGAN~\cite{isola2017image}, a fully convolutional network operating on patches with relatively fewer hyperparameters, from which we obtain the output of the discriminator in the form of a two-dimensional map. The loss for training $D_{img}$ can be written as follows:
\begin{align}
  \label{eq:loss_input}
  \cL_{input} = \mathbb{E}_{x_t \sim X_T}&\left[\log D_{img}(x_t)\right]\\
   + \mathbb{E}_{x_s \sim X_S}&\left[\log(1 - D_{img}(\hat{x}_s))\right],\nonumber
\end{align}
in which we omit the image width and height dimension for the sake of simplicity. 

As mentioned above, the transformed images are expected to be also useful for vision tasks at hand. This is achieved by joint training the image transform network with the task network (details are provided in the next section). Since the image transform network is differentiable, the gradients from the task network can guide the transform network to ensure the preservation of useful information from the synthetic data.

\subsection{Output-Level Adaptation}
Our task network $G_{task}$ concurrently performs depth estimation and semantic segmentation for a given input image. The network is shared between two domains and takes either a transformed synthetic image $\hat{x}_s$ or a real image $x_t$ as input. Specifically, a shared base feature extractor is built for the two tasks, with one head on top of it respectively for each task, namely one head for semantic segmentation output and the other one for depth estimation output. 

The semantic segmentation task is learned by minimizing a standard cross-entropy loss:
\begin{align}
  \label{eq:gan-loss}
  \cL_{seg} = \mathbb{E}_{x_s \sim X_S} [CE(y_s,\widetilde{y}_s)],
\end{align}
where ${y_s}$ stands for ground-truth semantic labels, and $\widetilde{y}_s$ stands for predicted labels. With regard to the semantic segmentation task, depth estimation can be seen as an auxiliary task. As a common practice, we employ the $\ell_1$ loss for the depth estimation task as follows:
\begin{align}
  \label{eq:gan-loss}
\cL_{depth} = \mathbb{E}_{x_s \sim X_S} [||d_s-\widetilde{d_s}||_1],
\end{align}
where ${d_s}$ stands for ground-truth depth, and $\widetilde{d_s}$ stands for predicted depth. Note that both losses only apply to the source domain, where the supervision is available. 

To ensure that the task network performs well in the target domain, we further apply a discriminator $D_{task}$ on the outputs of the tasks as inspired by \cite{tsai2018learning}. However, instead of using only the semantic space, our work jointly considers both semantics and depth, as the inherent correlation between semantics and depth information could be a helpful cue to effectively reduce domain difference. In particular, we concatenate the output of semantic segmentation map $\widetilde{y}_s$ (\resp $\widetilde{y}_t$) and the output of depth estimation map $\widetilde{d_s}$ (\resp $\widetilde{d}_t$), which leads to totally $C+1$ channels in the output maps. We use the concatenated maps to train the discriminator $D_{task}$ which distinguishes maps of the source domain from those of the target domain. Similar to $D_{img}$, $D_{task}$ is also formulated as a PatchGAN in favour of its awareness of spatial contextual relations. The loss for $D_{task}$ can be written as follows:
\begin{align}
  \label{eq:output_loss}
  \cL_{output} = \mathbb{E}_{x_t \sim X_T}&\left[\log D_{output}(\widetilde{d}_t,\widetilde{y}_t)\right]\\
   + \mathbb{E}_{x_s \sim X_S}&\left[\log(1 - D_{output}(\widetilde{d_s},\widetilde{y}_s))\right].\nonumber
\end{align}
\begin{table*}[t]
\center
\begin{tabular}{ c | c c c c c c c c c c | c }
\hline
& \vtext{road}& \vtext{building} & \vtext{pole} & \vtext{traffic light} & \vtext{traffic sign} & \vtext{vegetation}& \vtext{terrain} & \vtext{sky} & \vtext{car} & \vtext{truck} & \vtext{mIoU} \\ \hline
Non Adapt &
79.3 & 60.5 &  0.0 &  0.3 &  9.5 & 66.8 &  8.3 & 85.9 & 59.2 &  4.8 & 37.5 
\\ \hline

Input Level Adapt &
83.2 & 67.4 & 10.8 & 21.9 & 24.5 & 68.8 &  6.5 & 88.3 & 77.8 &  9.3 & 45.9 
\\ \hline

Output Level Adapt &
81.1 & 69.1 &  7.1 &  8.6 & 28.3 & 79.5 & 43.3 & 86.0 & 79.3 & 17.8 & 50.0 
\\ \hline

Input\&Ouput Adapt &
81.4 & 71.2 & 11.3 & 26.6 & 23.6 & 82.8 & 56.5 & 88.4 & 80.1 & 12.7 & \textbf{53.5} 
\\ \hline
\end{tabular}
\vspace{2mm}
\caption{\textbf{Experimental Results on Virtual KITTI$\rightarrow$KITTI.} The results are reported on mIoU over 10 categories. The best result is denoted in bold.}
\vspace{2mm}
\label{tab:vkitti}
\end{table*}

\subsection{Overall Training Objective}
We integrate both input-level and output-level modules, to train all networks $G_{task}$, $G_{img}$, $D_{img}$ and $D_{output}$ jointly. The overall objective is written as follows:
\begin{align}
    \label{eq:full_objective}
   \min_{\substack{G_{img}\\G_{task}}} & \max_{\substack{D_{img}\\ D_{output}}} \{\cL_{seg} +\lambda_{depth}\cL_{depth}\\ 
   + &\lambda_{image}\cL_{image}+\lambda_{output}\cL_{output}\}, \nonumber
\end{align}
where $\lambda$s act as the trade-off weights to balance different loss terms. The min-max problem is optimized with the adversarial training strategy. Note that domain adaptation procedure is only performed in the training phase. During test time, only $G_{task}$ is used on real images, and other components such as $G_{img}$, $D_{img}$ and $D_{output}$ can be removed.

\subsection{Implementation Details}
In our GIO-Ada model, the image transform network $G_{img}$ resembles the generator in CycleGAN~\cite{zhu2017unpaired}, which is based on the network in \cite{johnson2016perceptual} with several convolutional layers and residual blocks. For the task network, we employ the DeepLab-v2 model~\cite{chen2016deeplab}, due to its excellent performance and implementation availability. The VGG network~\cite{simonyan2014very} is used as the backbone model for the task network, which is initialized with the ImageNet pre-trained weights. Moreover, the discriminators are based on PatchGAN~\cite{isola2017image}, for which the weights are randomly initialized from a Gaussian distribution. 

The trade-off parameters are empirically set as $\lambda_{depth}=0.1,\lambda_{image}=0.1,\lambda_{output}=0.001$ in our experiments. We use Adam optimizer with initial learning rate $2^{-4}$. The network is trained for 10 epochs. Each mini-batch contains two images, one sampled from the source domain and the other sampled from the target domain. Random horizontal flip is used for data augmentation. 

\section{Experiments}
In this section, we verify the effectiveness of our proposed GIO-Ada model for semantic segmentation from synthetic data to real scenarios. 
\subsection{Experiment Settings}
Following the common unsupervised domain adaptation protocol, we use a synthetic dataset as the source domain, and a real dataset as the target domain. For the synthetic datasets, we employ Virtual KITTI~\cite{gaidon2016virtual} and SYNTHIA~\cite{ros2016synthia}, as depth information is publicly available for these two datasets. Accordingly, KITTI~\cite{geiger2013vision} and Cityscapes~\cite{cordts2016cityscapes} are used as the real datasets, which leads to two setting pairs: Virtual KITTI$\rightarrow$KITTI, and SYNTHIA$\rightarrow$Cityscapes. Other popular synthetic datasets such as GTA~\cite{richter2016playing} and  VIPER~\cite{richter2017playing} are not used in our experiments, since depth information is not provided for those datasets. We briefly introduce the datasets used in our experiments as below.

\paragraph{KITTI~\cite{geiger2013vision}} is a dataset focusing on autonomous driving, which consists of images depicting several driving urban scenarios. It is collected by moving vehicles in multiple cities. The official split for semantic segmentation is used in our experiment, which contains 200 training images, and 200 test images, with a spatial resolution around $1242\times 375$. Due to the ground-truth label is only available in training set, thus we use the official unlabelled test images to adapt our model, and we report the results on the official training set.

\paragraph{Virtual KITTI~\cite{gaidon2016virtual}} is a photo-realistic synthetic dataset which contains 21,260 images. Each image is densely annotated at pixel level with category and depth information. It is designed to mimic the conditions of KITTI dataset and has similar scene layout, camera viewpoint, and image resolution as KITTI dataset, thus making it ideal to study the domain adaptation problems between synthetic and real data. 

\paragraph{Cityscapes~\cite{cordts2016cityscapes}} consists of $2,975$ images in the training set, and $500$ images in the validation set. The images have a fixed spatial resolution of $2048\times 1024$ pixels. Due to the large size of image, as a common practice we down-size the images to half resolution (at $ 1024\times 512$). The training set is used to adapt the model, and we report our results on the validation set.

\paragraph{SYNTHIA~\cite{ros2016synthia}} is a dataset with synthetic images of urban scenes and pixel-wise annotations. The rendering covers a variety of environments and weather conditions. In our experiment, we adopt the SYNTHIA-RAND-CITYSCAPES subset, which contains 9,400 images compatible with the Cityscapes categories. 

\subsection{Results on Virtual KITTI$\rightarrow$KITTI}
We first evaluate the effectiveness of the proposed method for learning semantic segmentation from the Virtual KITTI dataset to the KITTI dataset. The 10 common categories between two datasets are used for performance evaluation. We summarize the mean of Intersection over Union (mIoU) in Table~\ref{tab:vkitti}. Overall, our GIO-Ada improves the mIoU over the non-adaptive baseline by $+16\%$, which confirms the effectiveness of our method for cross-domain semantic segmentation. To further study the benefits of the adaptation modules on different levels, we break down the performance by testing the ablated versions of our model: the input level adaptation achieves $+8.4\%$ performance gain, while the output level adaptation achieves $+12.5\%$ improvements. This demonstrates the effectiveness of both modules for adapting segmentation models form the synthetic domain to the real domain. Moreover, the two levels of adaptation modules are also shown to be complementary, as integrating them can further reduce the domain gap. 

\begin{figure*}[h]
\small
\centering

      \resizebox{\linewidth}{!}{%
	  \setlength{\fboxsep}{0pt}
      \fbox{\includegraphics[height=0.1\textwidth]{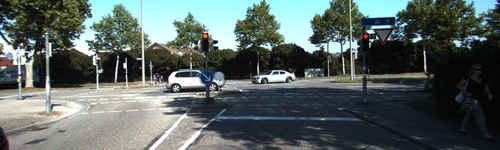}}
      \fbox{\includegraphics[height=0.1\textwidth]{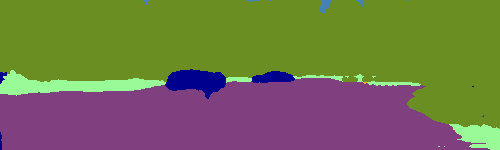}}
      \fbox{\includegraphics[height=0.1\textwidth]{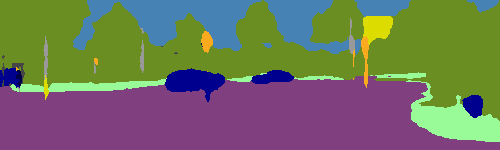}}
      }\\
      
      \resizebox{\linewidth}{!}{%
	  \setlength{\fboxsep}{0pt}
      \fbox{\includegraphics[height=0.1\textwidth]{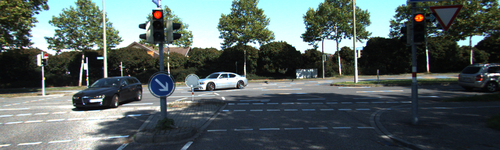}}
      \fbox{\includegraphics[height=0.1\textwidth]{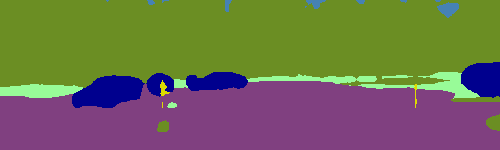}}
      \fbox{\includegraphics[height=0.1\textwidth]{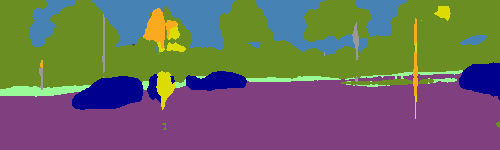}}
      }\\

      \resizebox{\linewidth}{!}{%
	  \setlength{\fboxsep}{0pt}
      \fbox{\includegraphics[height=0.1\textwidth]{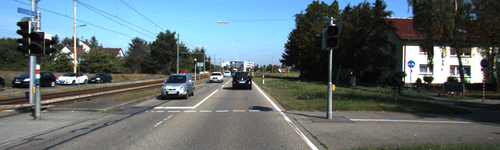}}
      \fbox{\includegraphics[height=0.1\textwidth]{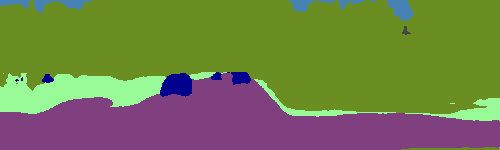}}
      \fbox{\includegraphics[height=0.1\textwidth]{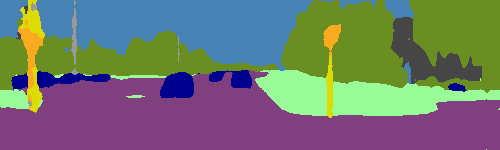}}
      }\\
      
      \resizebox{\linewidth}{!}{%
	  \setlength{\fboxsep}{0pt}
      \fbox{\includegraphics[height=0.1\textwidth]{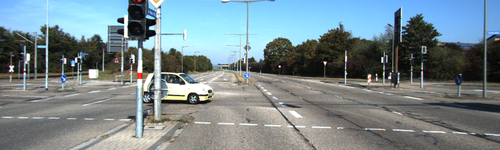}}
      \fbox{\includegraphics[height=0.1\textwidth]{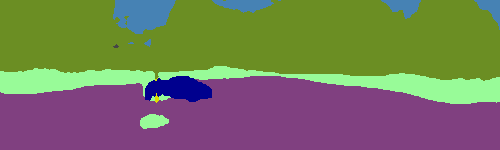}}
      \fbox{\includegraphics[height=0.1\textwidth]{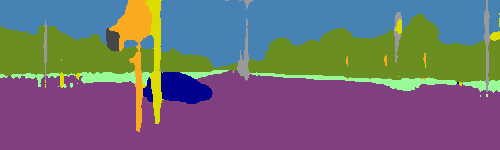}}
      }\\
      
      \resizebox{\linewidth}{!}{%
	  \setlength{\fboxsep}{0pt}
      \fbox{\includegraphics[height=0.1\textwidth]{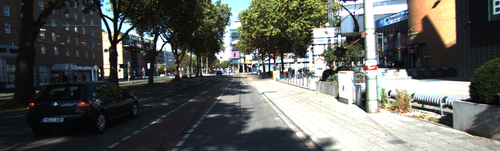}}
      \fbox{\includegraphics[height=0.1\textwidth]{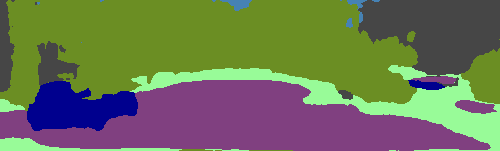}}
      \fbox{\includegraphics[height=0.1\textwidth]{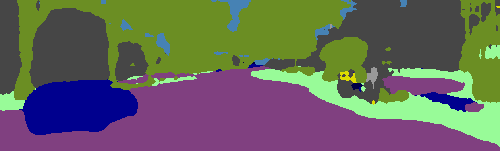}}
      }\\
      
      \vspace{5mm}
        \caption{\textbf{Semantic Segmentation Qualitative Results on KITTI dataset.} We follow the color encoding scheme of Cityscapes to colorize the label map. From left to right: \textbf{left:} input image, \textbf{middle:} non-adaptive results, and \textbf{right:} results by our method.}
        \vspace{5mm}
\label{fig:qualitative_kitti}
\end{figure*}

We also provide a few qualitative examples in Figure~\ref{fig:qualitative_kitti}. From those results, we observe that the segmentation results generally get improved with our GIO-Ada approach. Especially, by leveraging the geometric cues, our model produce excellent segmentation on objects with geometric structure, such as poles, traffic signs, etc., which are usually challenging for existing methods~\cite{chen2018road,hoffman2017cycada,hoffman2016fcns}. 

To further investigate the different design variants, especially with a focus on the importance geometric cue in the two components. We conduct further ablation studies on the two adaptation modules individually in below.

\paragraph{Ablation Study on Input Level Adaptation:}
\begin{table}[H]
\center
\begin{tabular}{ c c c c c c c }
\hline
\textbf{na} & \textbf{cg} & \textbf{gd} & \textbf{+d} & \textbf{+s} & \textbf{+sd} \\ \hline
37.5 & 39.8 & 43.5 & 44.2 & 44.7& \textbf{45.9} \\ \hline
\end{tabular}
\vspace{3mm}
\caption{\textbf{Ablation Study on Input Level Adaptation.} mIoU over 10 categories is reported. \textbf{na}: non-adaptive baseline; \textbf{cg}: image translation with CycleGAN~\cite{zhu2017unpaired}; \textbf{gd}:  image transform network is guided by the task network; \textbf{+d}: with additional depth input to the image transform network; \textbf{+s}: with additional semantic label input; \textbf{+sd}: with both depth and semantic labels as additional input, which is also our final model for input level adaptation.}
\vspace{3mm}
\label{tab:vkitti_input_ablation}
\end{table}

In our final input level adaptation model, we use an image transform network, which takes an image and its corresponding depth and semantic label as input. To investigate the benefits of using additional inputs, we tested three special cases of input level adaptation module with only depth, with only semantic label, or with none as additional input. We also include \cite{zhu2017unpaired}, an image translation model commonly adapted for domain adaptation, for comparison. 

The results are summarized in Table~\ref{tab:vkitti_input_ablation}. We observe that all other methods outperform the non-adaptive baseline, demonstrating the importance of input-level adaptation. However, CycleGAN only improves the baseline result by $+2.3\%$, which is less effective compared to the improvement of $+6\%$ achieved by the task network. This indicates that the gradient from the task network is a useful guidance for the image transform network to preserve useful information. Nevertheless, the performance can be further boosted when additional information is further taken as input. Adding individually depth and semantic segmentation as the additional input gives an improvement of $+6.7\%$ and $+7.2\%$, respectively, and integrating them together produces $+8.4\%$ performance gain. The results suggest that the geometric information can be very useful in the image transformation process in the sense that it helps to preserve rich information in the raw 3D model. 

\begin{figure*}[h]
\small
\centering

      \resizebox{\linewidth}{!}{%
	  \setlength{\fboxsep}{0pt}
      \fbox{\includegraphics[height=0.1\textwidth]{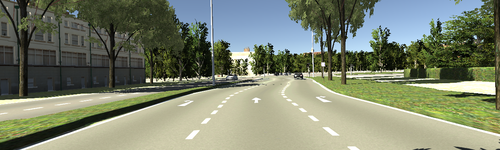}}
      \fbox{\includegraphics[height=0.1\textwidth]{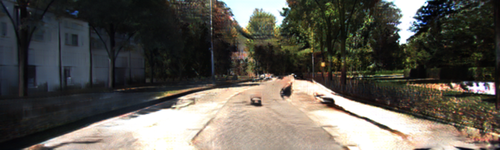}}
      \fbox{\includegraphics[height=0.1\textwidth]{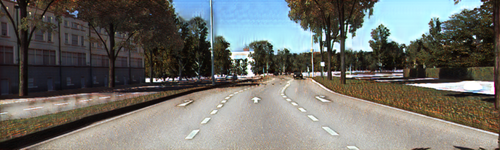}}
      }\\
     
      \resizebox{\linewidth}{!}{%
	  \setlength{\fboxsep}{0pt}
      \fbox{\includegraphics[height=0.1\textwidth]{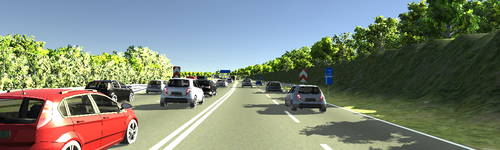}}
      \fbox{\includegraphics[height=0.1\textwidth]{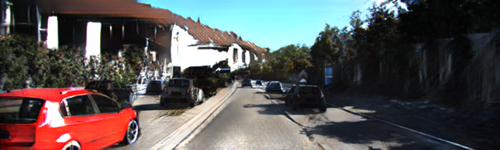}}
      \fbox{\includegraphics[height=0.1\textwidth]{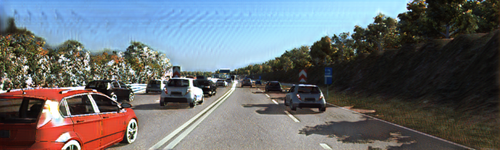}}
      }\\
      
      \resizebox{\linewidth}{!}{%
	  \setlength{\fboxsep}{0pt}
      \fbox{\includegraphics[height=0.1\textwidth]{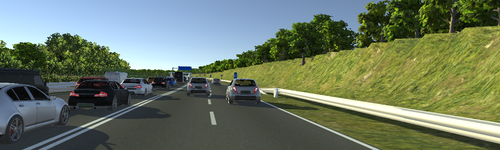}}
      \fbox{\includegraphics[height=0.1\textwidth]{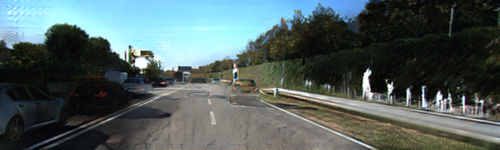}}
      \fbox{\includegraphics[height=0.1\textwidth]{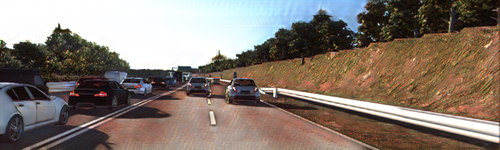}}
      }\\
     
      \resizebox{\linewidth}{!}{%
	  \setlength{\fboxsep}{0pt}
      \fbox{\includegraphics[height=0.1\textwidth]{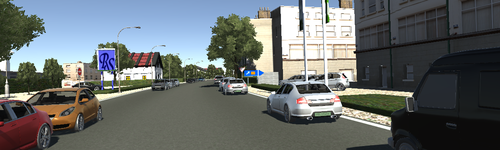}}
      \fbox{\includegraphics[height=0.1\textwidth]{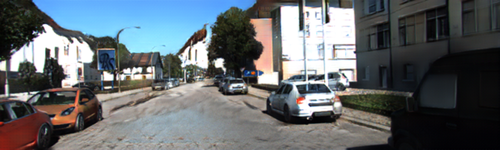}}
      \fbox{\includegraphics[height=0.1\textwidth]{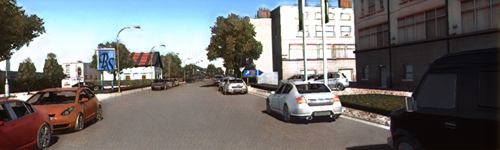}}
      }\\
      
      \resizebox{\linewidth}{!}{%
	  \setlength{\fboxsep}{0pt}
      \fbox{\includegraphics[height=0.1\textwidth]{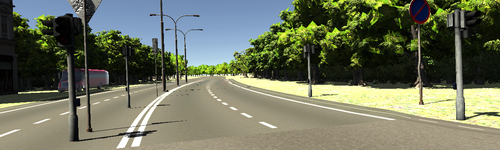}}
      \fbox{\includegraphics[height=0.1\textwidth]{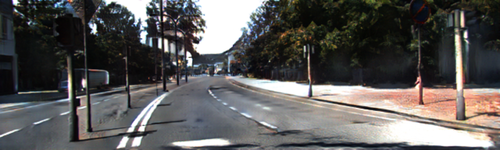}}
      \fbox{\includegraphics[height=0.1\textwidth]{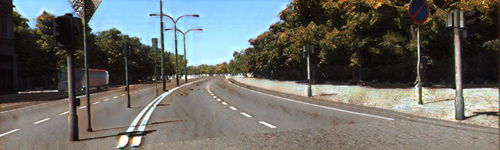}}
      }\\
      
      \vspace{5mm}
        \caption{\textbf{Qualitative Results on Input-level Adaptation.} From left to right: \textbf{left:} input synthetic image, we compare the image translation results of \textbf{middle:} CycleGAN, with \textbf{right:} our result.}
        \vspace{5mm}
\label{fig:qualitative_image_translation}
\end{figure*}

We further demonstrate this by providing a few examples of translated images with CycleGAN and our model in Figure~\ref{fig:qualitative_image_translation}, in which we clearly observe that our model is able to preserves more of the geometric and semantic consistency during the translation process. More specifically, CycleGAN are observed to hallucinate buildings and trees in the sky (row 1,2,4), the poles turn into trees (row 5), and cars turn to road (row 3). In comparison, our model is able to preserve the semantic and geometric consistency. 

\paragraph{Ablation Study on Output Level Adaptation:}

\begin{table}
\center
\begin{tabular}{c c c c c c c c c c}
\hline
\textbf{na} & \textbf{ss} & \textbf{depth} & \textbf{sep} & \textbf{joint} \\ \hline
37.5 & 45.9 & 43.8 & 46.3 & \textbf{50.0} \\ \hline
\end{tabular}
\vspace{2mm}
\caption{\textbf{Ablation Study on Output Level Adaptation.} mIoU over 10 categories is reported. \textbf{na}: the non-adaptive baseline; \textbf{ss}: aligning the semantic segmentation output map; \textbf{depth}: aligning the depth output map; \textbf{sep}: individually aligning both semantic segmentation and depth estimation; \textbf{joint}: aligning the joint output space of depth estimation and semantic segmentation, which is also our final output level adaptation model.}
\label{tab:vkitti_output_ablation}
\vspace{2mm}
\end{table}

\begin{table*}
\center
\resizebox{0.95\textwidth}{!}{%
\setlength{\tabcolsep}{3pt}
\begin{tabular}{ c | c c c c c c c c c c c c c c c c | c | c}
\hline

& \vtext{road}& \vtext{sidewalk} & \vtext{building} & \vtext{wall*} & \vtext{fence*} & \vtext{pole*} & \vtext{traffic light} & \vtext{traffic sign} & \vtext{vegetation} & \vtext{sky} & \vtext{person} & \vtext{rider} & \vtext{car} & \vtext{bus} & \vtext{motorbike} & \vtext{bicycle} & \vtext{mIoU} & \vtext{mIoU excl.*} \\ \hline 

FCNs Wld~\cite{hoffman2016fcns}& 
11.5 & 19.6 & 30.8 &  4.4 &  0.0 & 20.3 &  0.1 & 11.7 & 42.3 & 68.7 & 51.2 &  3.8 & 54.0 &  3.2 &  0.2 &  0.6 & 20.1 & 22.9 
\\ \hline

Curriculum~\cite{zhang2017curriculum}& 
65.2 & 26.1 & 74.9 &  0.1 &  0.5 & 10.7 &  3.7 &  3.0 & 76.1 & 70.6 & 47.1 &  8.2 & 43.2 & 20.7 &  0.7 & 13.1 & 29.0 & 34.8 
\\ \hline

Cross-City~\cite{chen2017no}& 
62.7 & 25.6 & 78.3 &    - &    - &    - &  1.2 &  5.4 & 81.3 & 81.0 & 37.4 &  6.4 & 63.5 & 16.1 &  1.2 &  4.6 &    - & 35.7 
\\ \hline

ROAD-Net~\cite{chen2018road}&
77.7 & 30.0 & 77.5 &  9.6 &  0.3 & 25.8 & 10.3 & 15.6 & 77.6 & 79.8 & 44.5 & 16.6 & 67.8 & 14.5 &  7.0 & 23.8 & 36.1 & 41.7 
\\ \hline

Tsai etal.~\cite{tsai2018learning}&
78.9 & 29.2 & 75.5 &    - &    - &    - &  0.1 &  4.8 & 72.6 & 76.7 & 43.4 &  8.8 & 71.1 & 16.0 &  3.6 &  8.4 &    - & 37.6 
\\ \hline

Sankaranarayanan etal.~\cite{sankaranarayanan2018learning}&
80.1 & 29.1 & 77.5 &  2.8 &  0.4 & 26.8 & 11.1 & 18.0 & 78.1 & 76.7 & 48.2 & 15.2 & 70.5 & 17.4 &  8.7 & 16.7 & 36.1 & 42.1 
\\ \hline

CBST~\cite{zou2018unsupervised}&
69.6 & 28.7 & 69.5 & 12.1 &  0.1 & 25.4 & 11.9 & 13.6 & 82.0 & 81.9 & 49.1 & 14.5 & 66.0 &  6.6 &  3.7 & 32.4 & 35.4 & 40.7 
\\ \hline

Ours (Non Adapt) &
 9.7 & 14.1 & 58.5 &  4.7 &  0.3 & 22.7 &  1.9 & 12.9 & 70.7 & 60.9 & 50.2 &  7.2 & 32.2 & 17.4 &  1.3 &  8.0 & 23.3 & 26.5 
 \\ \hline

Ours (Input-level Adapt) &
77.0 & 29.3 & 67.9 &  0.1 &  0.1 & 24.7 & 10.7 & 17.4 & 79.4 & 78.8 & 49.2 & 13.7 & 70.3 &  4.3 &  5.8 & 12.8 & 33.8 & 39.7 
\\ \hline

Ours (Output-level Adapt) &
79.6 & 29.7 & 75.7 & 11.4 &  0.3 & 25.3 & 11.1 & 14.8 & 76.7 & 76.9 & 45.3 & 15.9 & 67.7 & 15.8 &  4.8 & 13.5 & 35.3 & 40.6 
\\ \hline

Ours (Input\&Output Adapt) &
78.3 & 29.2 & 76.9 & 11.4 &  0.3 & 26.5 & 10.8 & 17.2 & 81.7 & 81.9 & 45.8 & 15.4 & 68.0 & 15.9 &  7.5 & 30.4 & \textbf{37.3} & \textbf{43.0}
\\ \hline

\end{tabular}
}
\vspace{3mm}
\caption{\textbf{Comparison with state-of-the-arts methods for semantic segmentation on Cityscapes adapting from SYNTHIA.} Results of state-of-the-art methods are collected from the original papers. All results are based on VGG as the backbone architecture. Some works only evaluate on 13 classes, we hereby mark these excluded categories with *. We also report the average performance over 13 classes as \textit{mIoU excl. *}. The best results are denoted in bold.}
\vspace{3mm}
\label{tab:synthia}
\end{table*}

We also study different variants of the output-level adaptation. There are several possible alternatives to our joint output space adaptation. For example, performing the output space alignment proposed by~\cite{tsai2018learning} in semantic segmentation space and depth estimation space separately. Additionally, we try to build two discriminators to individually align the two output spaces, without considering the correlation between the two tasks. We compare these variants to our final model which aligns the joint space of depth and semantic segmentation outputs.

The results are shown in Table~\ref{tab:vkitti_output_ablation}. First, we observe that all variants achieve significant improvement over the baseline. This shows the effectiveness of domain adaptation techniques in general. Particularly, output space alignment on semantic segmentation prediction~\cite{tsai2018learning} achieves performance gain of $+8.4\%$, while the improvement of the same output space adaptation module on depth prediction is $+6.3\%$. This is not surprising, considering our final objective is semantic segmentation. Aligning the semantic segmentation map would have a more direct influence on the segmentation results. We then combine depth alignment and semantic segmentation alignment, which gives an  improvement of $+8.8\%$ over baseline, marginally better than using only semantic segmentation alignment. This suggests that trivially optimizing each task can not bring in performance gain without modeling the correlation between the tasks. Finally, by aligning the joint space of depth and semantic segmentation, we achieve a significant improvement of $+12.5\%$,  showing that the key to reducing domain shift is to use joint correlation of the two outputs, which also verifies our motivations.

\subsection{Results on SYNTHIA$\rightarrow$Cityscapes}
To facilitate the comparison with other state-of-the-art works, we further evaluate the proposed method on SYNTHIA to Cityscapes setting following \cite{hoffman2016fcns,zhang2017curriculum,chen2017no,chen2018road,tsai2018learning,sankaranarayanan2018learning,zou2018unsupervised}. 
The results of all methods are summarized in Table~\ref{tab:synthia}. For a fair comparison, all methods used VGG-16 as the backbone network. 

Similarly to the setting of Virtual KITTI $\rightarrow$ KITTI, the adaptation at both input and output levels is helpful for performance improvement: the input-level adaptation improves the baseline by $+10.5\%$, while the output-level adaptation improves it by $+12.0\%$. Integrating the two modules gives a larger performance gain of $+14.0\%$ over the non-adaptive baseline. This again verifies the effectiveness of our adaptation modules in both the input and output levels. 

Our model outperforms all other competing methods by a healthy margin. We attribute this to the supplement of geometric cues to the semantic segmentation task during domain adaptation.

\section{Conclusion}
In this paper, we have introduced a new \textit{Geometrically Guided Input-Output Adaptation} (GIO-Ada) model, which effectively leverages the costless geometric information in synthetic data to tackle the cross-domain semantic segmentation problem. Geometrically guided adaptation is performed on two different levels: 1) on the input level, depth information together with the semantic annotation is used as additional input for guiding the image transform network to reduce the domain shift on raw pixels, and 2) on the output level, depth prediction and semantic prediction are used to form a joint output space, on which an adversarial training strategy is applied to reduce the domain shift. We have experimentally validated our method on two pairs of datasets. The results demonstrate effectiveness of our GIO-Ada for cross-domain semantic segmentation with leveraged geometric information from virtual data.

\newpage
{\small
\bibliographystyle{ieee}
\bibliography{main_bib}
}

\end{document}